\documentclass[pdflatex,sn-mathphys-num]{sn-jnl}

\usepackage{graphicx}
\usepackage{multirow}
\usepackage{amsmath,amssymb,amsfonts}
\usepackage{amsthm}
\usepackage{mathrsfs}
\usepackage[title]{appendix}
\usepackage{xcolor}
\usepackage{textcomp}
\usepackage{manyfoot}
\usepackage{booktabs}
\usepackage{array}
\usepackage{tabularx}
\usepackage{url}
\usepackage{float}
\usepackage{placeins}
\providecommand{\doiurl}[1]{%
	\href{https://doi.org/#1}{\nolinkurl{https://doi.org/#1}}%
}
\begin{document}

\title[Lightweight Global-Target KAN for UAV Imagery]{A Lightweight Global-Target Framework for Multi-Domain No-Reference Image Quality Assessment in UAV Imagery}

\author*[1]{\fnm{Koffi Titus Sergio} \sur{Aglin}}\email{sergioaglin85@gmail.com}
\author[2]{\fnm{Anthony K.} \sur{Muchiri}}
\author[3]{\fnm{Celestin} \sur{Nkundineza}}

\affil*[1]{\orgdiv{Department of Mechatronic Engineering}, \orgname{Institute for Basic Sciences, Technology and Innovation, Pan African University (PAUSTI)}, \orgaddress{\city{Nairobi}, \postcode{00100}, \country{Kenya}}}
\affil[2]{\orgdiv{Department of Mechatronic Engineering}, \orgname{Jomo Kenyatta University of Agriculture and Technology (JKUAT)}, \orgaddress{\city{Nairobi}, \postcode{00200}, \country{Kenya}}}
\affil[3]{\orgname{College of Science and Technology, University of Rwanda}, \orgaddress{\city{Kigali}, \country{Rwanda}}}

\abstract{Reliable image-quality assessment is important in Unmanned Aerial Vehicle surveys because degraded frames can compromise mapping and downstream visual analysis. This paper presents Global-Target Assessment and Regression (GloTAR), a lightweight multi-domain No-Reference Image Quality Assessment framework combining whole-image context with saliency-derived regional evidence and Kolmogorov--Arnold Network layers for compact feature reduction and regression. The model is trained with balanced sampling across five public IQA datasets containing authentic and synthetic degradations. On 6,215 held-out images, the primary evaluation achieves macro SRCC 0.8050 and PLCC 0.8271. Across five independent training runs with fixed data partitions, the full model obtains macro SRCC {\mathversion{normal}$0.8124\pm0.0062$} and PLCC {\mathversion{normal}$0.8358\pm0.0065$}. On Drone-IQA, the corresponding five-run SRCC is {\mathversion{normal}$0.8933\pm0.0023$} and PLCC is {\mathversion{normal}$0.8987\pm0.0018$}. The network has 6.525 million trainable parameters and reaches 15.58~ms model-only latency and 64.19 frames per second on an NVIDIA Tesla T4 GPU.}


\maketitle

\section{Introduction}\label{sec:intro}
	
	Unmanned aerial systems have become an important source of high-spatial-resolution imagery for mapping, inspection, environmental monitoring \cite{cuaran2021crop}, and agricultural observation \cite{wang2024uas}. Small UAVs can operate at low altitude and revisit areas frequently, but the resulting imagery is vulnerable to platform motion, vibration, changing illumination, atmospheric effects, defocus, compression, and sensor noise. Motion blur \cite{li2023motionblur}, for example, can reduce the visibility of fine scene structures. When degraded frames are accepted without screening, errors can propagate into orthomosaic generation, detection, classification, and quantitative survey products.
	
	Image quality assessment (IQA) provides a mechanism to score or reject such frames. In field acquisition, however, a pristine reference image is generally unavailable; this motivates blind or no-reference image quality assessment (NR-IQA). Modern deep NR-IQA methods achieve high correlations with subjective opinion scores, but many rely on relatively large CNN, Transformer, or vision-language backbones, as reviewed in Section~\ref{sec:related}. This creates a practical tension between perceptual accuracy and the latency and memory budgets of UAV-oriented or near-edge processing pipelines. Embedded deployment is an important target use case, but it must ultimately be established on the intended hardware rather than inferred from a datacenter GPU benchmark.
	
	The present work addresses this trade-off with a compact dual-branch architecture. Rather than modeling UAV images as an isolated domain, the model is trained across both authentic and synthetically distorted public IQA datasets and is then specialized through an explicit saliency-region branch and UAV-specific evaluation. This choice is deliberate: a useful UAV quality model should remain sensitive to generic degradations while retaining information from locally salient regions that may matter for later analysis.
	
	The proposed model contains two independent MobileNetV3-Small encoders \cite{howard2019mobilenetv3}. The first processes the whole image to preserve global context; the second aggregates up to six salient or target regions. Compact Kolmogorov--Arnold Network (KAN) heads \cite{liu2024kan} reduce the branch representations and perform nonlinear fusion. This use of global and salient-region information differs from lightweight NR-IQA designs such as LAR-IQA \cite{avanaki2024lariqa}, which also explores MobileNetV3 and KAN regression but uses synthetic/authentic distortion branches rather than a global-target decomposition within one fused predictor. The methodological contribution is the explicit separation and subsequent fusion of whole-image and saliency-derived regional quality evidence in a compact multi-domain predictor, rather than a new backbone, saliency detector, or regression family. The deterministic spectral-residual proposals do not require target annotations; accordingly, the target branch processes salient regions that need not coincide with annotated Drone-IQA targets.

The principal contributions of this study are the development of a lightweight global--regional NR-IQA architecture for multi-domain image-quality prediction with a UAV-oriented evaluation, and the investigation of its accuracy--computational-efficiency trade-off under a controlled GPU benchmark.

	\section{Related work}\label{sec:related}
	
	\subsection{Deep no-reference image quality assessment}
	
	NR-IQA has moved from hand-crafted natural-scene statistics to learned representations with increasingly strong semantic modeling. BRISQUE\cite{mittal2012brisque}, for example, predicts quality from deviations in locally normalized luminance statistics without requiring a reference image. Early deep approaches then learned the quality representation directly: BIECON\cite{kim2017biecon} used a CNN trained with local quality guidance, while MEON\cite{ma2018meon} coupled distortion identification and quality prediction in a multi-task network. These methods remain useful reference points because they show how much accuracy can be gained before resorting to today's larger architectures. More recent models push performance further. HyperIQA\cite{su2020hyperiqa} uses a content-adaptive hypernetwork; MUSIQ\cite{ke2021musiq} models image quality across multiple scales with a Transformer; and TReS\cite{golestaneh2022tres} combines CNN features, Transformer modeling, relative ranking, and self-consistency. DEIQT improves data efficiency with an attention-panel decoder \cite{qin2023deiqt}, whereas LoDa\cite{xu2024loda}, TOPIQ\cite{chen2024topiq}, QualiCLIP\cite{agnolucci2024qualiclip}, and PerceptCLIP\cite{zalcher2025perceptclip} exploit large pretrained visual or vision-language representations.
	These methods demonstrate that semantic and long-range information can improve quality prediction, but the associated backbones may be expensive for continuous resource-constrained analysis. Lightweight NR-IQA is therefore not simply a model-compression problem; the representation itself must preserve quality-relevant evidence. LAR-IQA addresses this issue with mobile encoders, multiple color spaces, and KAN regression; its reported KAN configuration uses about 21.1 M parameters and obtains SRCC 0.803 on the UHD-IQA validation protocol \cite{avanaki2024lariqa}. A recent compact CNN BIQA model uses approximately 0.9 M parameters and reports SRCC/PLCC 0.798/0.789 on KADID-10K and 0.846/0.871 on TID2013. The present method is complementary. It separates global scene quality from saliency-derived regional quality and fuses both representations in a compact KAN head. Recent representation-learning studies in adjacent imaging and sensing
	problems provide additional methodological context. WaterCycleDiffusion
	combines visual--textual fusion with diffusion-based modeling for underwater
	image enhancement \cite{wang2026watercycle}. A visual--textual mutual-guidance
	fusion network integrates image and language representations for remote-sensing
	visual question answering \cite{liu2026vmgn}. Graph representation learning has
	also been applied to plot-to-track association in compact high-frequency
	surface-wave radar \cite{li2026plottrack}. For adaptive underwater image
	enhancement, a self-organized framework selects and organizes enhancement
	operations according to image degradation characteristics
	\cite{wang2024selforganized}. The Meta Underwater Camera similarly formulates
	underwater enhancement as an adaptive protocol for correcting color,
	contrast, and detail degradation \cite{wang2023meta}.
	
	\subsection{UAV-oriented image quality assessment}
	
	UAV image quality has received substantially less attention than general-purpose IQA. Ieremeiev et al. combined multiple no-reference metrics through an artificial neural network and evaluated the approach on TID2013 and real UAV images with embedded distortions \cite{ieremeiev2023uav}. Tian et al. developed a machine-learning NR-IQA method for UAV hyperspectral images containing blur, noise, stripe noise, and compound distortions; their best Extra Trees configuration obtained PLCC 0.963 and SROCC 0.925 on their dedicated UAV-HSI protocol \cite{tian2024uavhsi}. Lin et al. proposed a reference-free Swin-Unet UAV screening method with approximately 0.3 s per-image inference and RMSE 0.04 under their protocol \cite{lin2026uaviqa}. The ICME 2026 Drone-IQA Grand Challenge further formalized target-aware low-altitude UAV IQA with global, target, and background annotations and PLCC/SRCC evaluation \cite{jiang2026droneiqa}. DroneIQA-VLE, a challenge solution, ensembles SigLIP2 encoders with a LoRA-adapted Qwen3.5-9B vision-language model and reported second place \cite{sun2026droneiqavle}.
	
	These UAV studies are not numerically interchangeable because they use different modalities, labels, splits, and evaluation protocols. They nevertheless establish two practical requirements that motivate the present work: reference-free quality scoring and computational feasibility during high-frequency image acquisition.
	
	\section{Materials and methods}\label{sec:methods}
	
	\subsection{Problem formulation}
	This study proposes the Global-Target Assessment and Regression (GloTAR) framework, a lightweight multi-domain no-reference image quality assessment (NR-IQA) approach designed to estimate perceptual image quality while jointly exploiting global image context and salient-region information. GloTAR consists of two complementary processing branches: a global branch that analyzes the complete image and a regional branch that analyzes salient regions extracted from the image. The resulting representations are compacted and fused using Kolmogorov–Arnold Network (KAN) layers to produce a single predicted quality score. The framework is designed for multi-domain training while maintaining particular relevance to computationally constrained UAV image-quality assessment. The name ``Global-Target'' is retained as the model identifier; operationally, however, the second branch is saliency-derived and is not claimed to correspond to annotated semantic targets.
	
	For an observed RGB image $I$, where $I$ denotes the complete input image, 
	NR-IQA estimates a scalar perceptual quality score without access to a pristine 
	reference. All dataset-specific opinion scores are mapped to a common normalized 
	quality variable $Q \in [0,1]$, with larger values indicating better visual quality. 
	GloTAR uses the complete image together with six target-branch input regions $R_k$, 
	where $R_k$ denotes the $k$-th salient region extracted from $I$, and 
	$k \in \{1,\ldots,6\}$ indexes the six target-branch input slots:
	\begin{equation}
		\widehat{Q} = F\left(I,\{R_k\}_{k=1}^{6}\right).
	\end{equation}
	Here, $F$ denotes the proposed GloTAR prediction function and $\widehat{Q}$ is the 
	predicted normalized image-quality score. The six target-region slots are produced 
	by the deterministic saliency-based proposal stage described below. The same 
	procedure is applied to every dataset, so target-region generation does not depend 
	on object bounding boxes or target annotations.
	
	\subsection{Target-region generation and preprocessing}
	
	Target proposals are generated with the spectral-residual saliency method of Hou and Zhang \cite{hou2007saliency}, implemented directly in the released code. Each RGB image is converted to grayscale and resized to a width of 64 pixels while preserving its aspect ratio, with a minimum working height of 32 pixels. A two-dimensional Fourier transform is computed, and the spectral residual is obtained by subtracting a $3\times3$ local average from the log-amplitude spectrum. The residual is combined with the original phase, transformed back to the spatial domain, squared, smoothed with a $5\times5$ Gaussian filter, and resized to the original image dimensions.
	
	The resulting saliency map is normalized to $[0,1]$ and binarized with Otsu thresholding. External connected contours define candidate regions. Bounding boxes covering less than 0.5\% of the image area are discarded; the remaining boxes are scored by their mean saliency and expanded by 10\% of their width and height before clipping to the image boundary. Candidates are sorted by saliency score and the six highest-ranked valid boxes are retained. If no valid proposal is found, the full image is used as the first region. When fewer than six regions remain, the unused target slots are filled with zero tensors. Consequently, the target branch always receives exactly six tensors and Eq.~(2) performs a fixed six-slot mean rather than a variable-$K$ mean. This fixed denominator is retained even when some slots are zero-padded; the method does not renormalize by the number of detected regions. The proposals are image-saliency regions, not ground-truth object detections, and may include background content.

	Input preprocessing is deliberately simple. Global images are resized to $384\times384$, while each selected region is resized to $224\times224$. Both branches are converted to tensors and normalized with ImageNet mean and standard-deviation values. No stochastic geometric or photometric augmentation is applied in the reported training run; this avoids introducing transformations whose effect on perceptual quality would not be reflected in the supplied opinion score.
	
	\subsection{Global-Target Assessment and Regression (GloTAR) architecture}
	
	Figure~\ref{fig:architecture} summarizes the proposed GloTAR architecture. 
	The global branch receives a $384\times384$ representation of the complete 
	RGB image $I$. A MobileNetV3-Small backbone extracts a compact feature 
	representation, which is passed through a KAN downsampler to produce a 
	128-dimensional global embedding $\mathbf{f}_g\in\mathbb{R}^{128}$, where 
	$\mathbf{f}_g$ represents the learned global image-quality features.
	
	The target branch receives the six $224\times224$ target-region tensors 
	$R_k$ described above. An independent MobileNetV3-Small encoder is used 
	to extract features from each region. Let $\phi_t(\cdot)$ denote the 
	target-branch feature extraction and embedding function. The six region 
	embeddings are computed independently and mean-pooled as
	\begin{equation}
		\mathbf{f}_t=
		\frac{1}{6}\sum_{k=1}^{6}\phi_t(R_k),
		\qquad
		\mathbf{f}_t\in\mathbb{R}^{128},
	\end{equation}
	where $\mathbf{f}_t$ denotes the resulting 128-dimensional target-region 
	feature embedding.
	
	The global and target embeddings are then concatenated to form the fused 
	feature representation
	\begin{equation}
		\mathbf{f}=
		[\mathbf{f}_g;\mathbf{f}_t]
		\in\mathbb{R}^{256},
	\end{equation}
	where $\mathbf{f}$ denotes the 256-dimensional joint representation 
	containing both global and target-region information. This representation 
	is passed to a KAN fusion head with hidden widths of 256 and 128 before 
	scalar quality regression. The two branch embeddings are concatenated channel-wise. Region features are aggregated before fusion, so the final head receives one 128-dimensional vector from each branch and outputs one scalar for the image. KAN layers replace fixed node-wise 
	nonlinearities with learnable univariate edge functions, offering a 
	flexible alternative to conventional multilayer perceptrons 
	\cite{liu2024kan}.
	
	\par\medskip
	\noindent\begin{minipage}{\linewidth}
		\centering
		\refstepcounter{figure}\label{fig:architecture}
		\includegraphics[width=\linewidth]{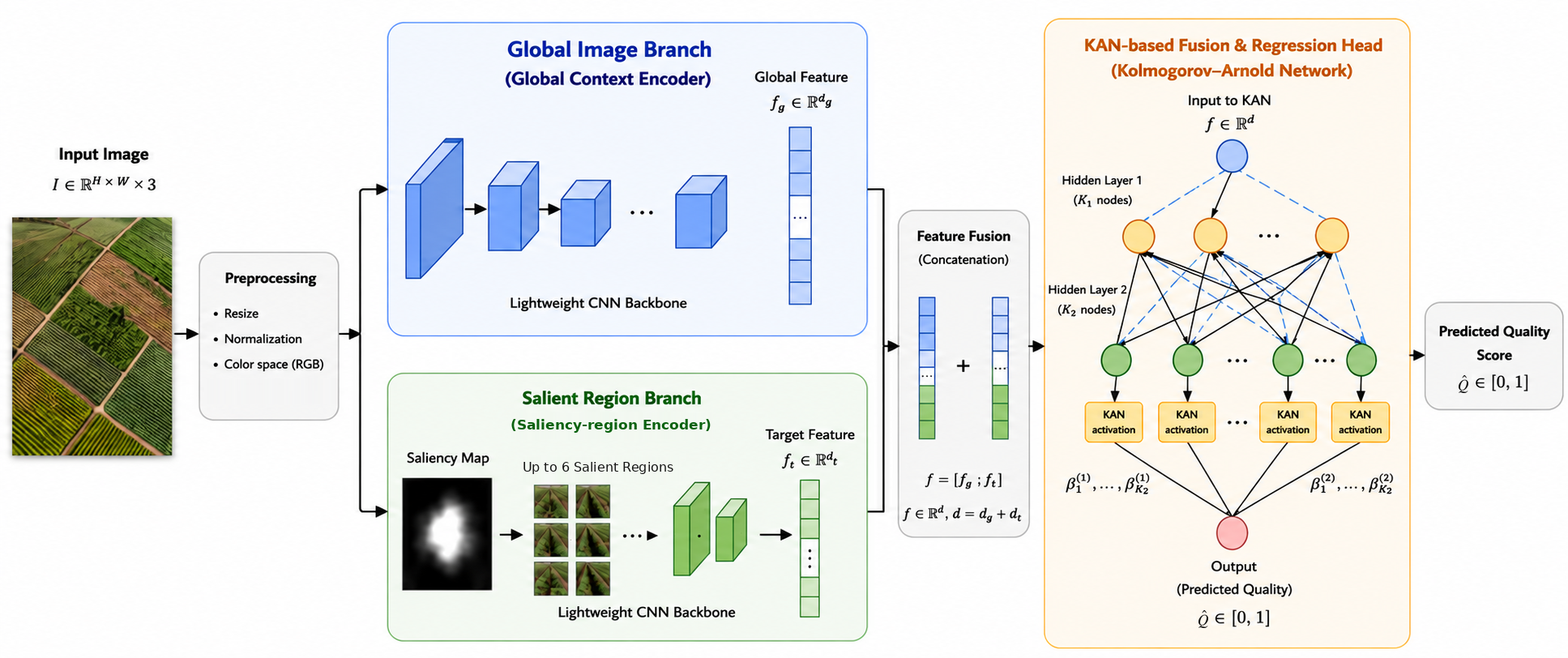}
		\par\smallskip
		{\sffamily\small\textbf{Figure~\thefigure:} Proposed global-target KAN architecture.}
	\end{minipage}
	\par\medskip
	
	\subsection{Training configuration}
	
	Both MobileNetV3-Small encoders are initialized from ImageNet-pretrained weights. The backbone produces a 576-dimensional feature vector for each input. In each branch, this representation is reduced by a KAN downsampler with dimensions $576\rightarrow256\rightarrow128$; the concatenated 256-dimensional global--regional representation is processed by a KAN regression head with dimensions $256\rightarrow256\rightarrow128\rightarrow1$. The KAN modules use the default configuration of the installed \texttt{efficient-kan} implementation: grid size 5, spline order 3, spline-noise scale 0.1, base scale 1.0, spline scale 1.0, SiLU base activation, grid adaptation parameter 0.02, and grid range $[-1,1]$. Training uses batch size 32, AdamW with an initial learning rate of $5\times10^{-5}$ and weight decay $10^{-4}$, and cosine annealing over a maximum of 20 epochs. The objective is mean squared error (MSE) between predicted and ground-truth unified MOS, with gradient clipping at a maximum norm of 5. Dataset-balanced sampling assigns each image from dataset $d$ the weight $1/n_d$, where $n_d$ is the number of training images in that dataset. Sampling is performed with replacement, with \texttt{num\_samples} equal to the full training-set size; consequently, each epoch contains 27,983 sampled instances. After every epoch, SRCC is computed separately on each validation dataset and macro-averaged. The checkpoint with the highest validation macro-SRCC is retained, and the held-out test set is evaluated only after model selection. The complete model contains 6,525,376 trainable parameters and occupies approximately 24.89 MB in FP32 parameter storage.

Subjective scores from the source datasets are expressed in their original
higher-is-better quality direction. The released manifest stores both each image's source-domain score
(\texttt{native\_score}) and the corresponding normalized value
(\texttt{quality\_score}), making the transformation auditable image by image.
Because the source datasets use different numerical score ranges, the scores
are affinely normalized to a common quality scale $Q\in[0,1]$, while preserving
their original ordering. The regression target is then expressed on the unified
scale $\mathrm{MOS}=1+4Q$. Drone-IQA is already on the 1--5 higher-is-better scale, for which $Q=(\mathrm{MOS}-1)/4$. For each source, the fixed affine score conversion and its direction are encoded in the released manifest through the \texttt{native\_score} and \texttt{quality\_score} fields; in particular, distortion-severity scores are reversed when larger native values indicate poorer quality. The conversion is applied consistently across training, validation, and test subsets, without fitting a new mapping to the held-out scores. The training set combines authentic and synthetic IQA sources rather than fitting only UAV imagery. This multi-domain design is intended to expose the predictor to diverse distortion statistics while retaining a UAV-oriented saliency-region pathway.
	
	\subsection{Datasets and held-out evaluation}
	
	Five datasets are used. TID2013\cite{ponomarenko2015tid} contains 3,000 synthetically distorted images generated from 25 references with 24 distortion types and five levels. KADID-10K\cite{lin2019kadid} contains 10,125 distorted images generated from 81 references across 25 distortions and five levels. KonIQ-10K\cite{hosu2020koniq} provides 10,073 authentically distorted, crowd-rated images captured in the wild. BIQ2021\cite{ahmed2022biq} is a large blind-IQA database with naturally occurring distortions. Drone-IQA\cite{jiang2026droneiqa} is a target-aware low-altitude UAV benchmark whose public training data include global quality and optional target/background annotations.
	
	The fixed train/validation/test partition contains 27,983/4,600/6,215 images, respectively. The held-out test set comprises 1,800 BIQ2021 images, 540 Drone-IQA images, 1,500 KADID-10K images, 2,015 KonIQ-10K images, and 360 TID2013 images. For the reference-based synthetic datasets, partitioning is reference-disjoint: TID2013 uses 19/3/3 reference images for train/validation/test, and KADID-10K uses 57/12/12 references, with no reference identity shared across partitions. This prevents distorted versions derived from the same pristine reference from appearing on both sides of an evaluation boundary. For the remaining authentic-image datasets, the evaluation uses the fixed image-level split identifiers supplied with the reproducibility manifest; the split is held unchanged across all reported random-seed runs. These image-level partitions are distinct from the reference-disjoint protocol used for the synthetic datasets.

For Drone-IQA, the internal partition contains 2,520 training, 540 validation, and 540 test images (70/15/15\%). The MOS mean $\pm$ standard deviation is $3.1675\pm0.6855$ for training, $3.1045\pm0.7123$ for validation, and $3.1450\pm0.7070$ for testing. The observed MOS ranges are 1.2222--4.6111, 1.1667--4.5000, and 1.1111--4.5556, respectively.
	
	\subsection{Evaluation metrics}
	
	The performance of the model is evaluated using correlation- and error-based metrics. Rank monotonicity is assessed using Spearman's rank correlation coefficient (SRCC) while linear agreement between predicted and ground-truth quality scores is measured using Pearson's linear correlation coefficient (PLCC).
	
	SRCC is defined as
	\begin{equation}
		\mathrm{SRCC}
		=
		1-\frac{6\sum_{i=1}^{N}d_i^2}
		{N(N^2-1)},
	\end{equation}
	where $d_i$ denotes the difference between the ranks of the ground-truth and predicted quality scores for the $i$th image, and $N$ is the number of images.
	
	PLCC is computed as
	\begin{equation}
		\mathrm{PLCC}
		=
		\frac{
			\sum_{i=1}^{N}(Q_i-\bar{Q})(\widehat{Q}_i-\overline{\widehat{Q}})
		}{
			\sqrt{
				\sum_{i=1}^{N}(Q_i-\bar{Q})^2
				\sum_{i=1}^{N}(\widehat{Q}_i-\overline{\widehat{Q}})^2
			}
		},
	\end{equation}
	where $Q_i$ and $\widehat{Q}_i$ denote the ground-truth and predicted quality scores, respectively, while $\bar{Q}$ and $\overline{\widehat{Q}}$ are their corresponding mean values.
	
	All four metrics are computed directly from the model predictions and ground-truth scores; no logistic or other nonlinear post-hoc mapping is fitted before PLCC, RMSE, or MAE computation.

	Prediction error on the common normalized quality scale is measured using root mean squared error (RMSE) and mean absolute error (MAE):
	\begin{align}
		\mathrm{RMSE}
		&=
		\sqrt{\frac{1}{N}
			\sum_{i=1}^{N}(Q_i-\widehat{Q}_i)^2},\\
		\mathrm{MAE}
		&=
		\frac{1}{N}
		\sum_{i=1}^{N}|Q_i-\widehat{Q}_i|.
	\end{align}

	\subsection{Controlled computational benchmark}
	
	Computational efficiency is evaluated on an NVIDIA Tesla T4 GPU with FP32 inference and batch size 1. Six deep NR-IQA baselines are loaded through the same PyIQA environment: HyperIQA \cite{su2020hyperiqa}, MUSIQ \cite{ke2021musiq}, TReS \cite{golestaneh2022tres}, MANIQA \cite{yang2022maniqa}, TOPIQ-NR \cite{chen2024topiq}, and QualiCLIP \cite{agnolucci2024qualiclip}. Baselines are timed on standardized $1\times3\times512\times512$ tensors after warm-up, with CUDA synchronization around each timed forward pass. GloTAR is timed with its native $384\times384$ global tensor and six $224\times224$ target tensors.
	
	Model-only latency: 15.58 ms per image, measures the GPU neural-network forward pass with prepared global and region tensors; it excludes image decoding, spectral-residual saliency extraction, region proposal and cropping, and input preparation. For GloTAR, a second benchmark measures the released deterministic predictor pipeline, including image loading, resizing, spectral-residual target proposal, crop formation, tensor preparation, and model execution, yielding 27.40 ms per image. The difference between these measurements is an aggregate pipeline overhead, not an independently measured saliency-only time.
	
	FLOPs are estimated with \texttt{fvcore}. The profiler counts 0.5301 GFLOPs for GloTAR, but several elementwise and KAN operations are unsupported by the tracer. We therefore report this column explicitly as \emph{fvcore-counted GFLOPs}, not exact total operations. The same caveat applies to unsupported operators in the profiled baseline architectures.
	
\section{Results}\label{sec:results}

This section presents the experimental results obtained with the proposed GloTAR framework, including its multi-domain prediction performance, comparison with representative NR-IQA methods, computational efficiency, and UAV-specific performance.

\subsection{Held-out multi-domain performance}

Table~\ref{tab:finalresults} reports the final held-out performance. Across all five datasets, GloTAR obtains macro SRCC 0.8050, PLCC 0.8271, RMSE 0.1151, and MAE 0.0911. The highest held-out correlation values occur on Drone-IQA, where SRCC reaches 0.9016 and PLCC reaches 0.9002. KADID-10K is next in rank correlation (SRCC 0.8402), while BIQ2021 is the most challenging held-out domain (SRCC 0.7210).

\begin{table}[htbp]
	\caption{Final held-out performance of GloTAR.}
	\label{tab:finalresults}
	\centering
	\small
	\begin{tabular*}{\linewidth}{@{\extracolsep{\fill}}lrrrrr@{}}
		\toprule
		Dataset & $N$ & SRCC & PLCC & RMSE & MAE\\
		\midrule
		BIQ2021
		& 1800 & 0.7210 & 0.7617 & 0.1186 & 0.0927\\
		
		Drone-IQA
		& 540 & \textbf{0.9016} & \textbf{0.9002}
		& \textbf{0.0874} & \textbf{0.0685}\\
		
		KADID-10K
		& 1500 & 0.8402 & 0.8327 & 0.1515 & 0.1203\\
		
		KonIQ-10K
		& 2015 & 0.7880 & 0.8238 & 0.0945 & 0.0722\\
		
		TID2013
		& 360 & 0.7743 & 0.8172 & 0.1232 & 0.1018\\
		
		\midrule
		
		Macro average
		& 6215 & \textbf{0.8050} & \textbf{0.8271}
		& \textbf{0.1151} & \textbf{0.0911}\\
		
		\bottomrule
	\end{tabular*}
\end{table}

\subsection{Training stability across random seeds}\label{sec:multiseed}

To assess sensitivity to training stochasticity, the complete GloTAR model was independently trained with five random seeds while keeping the train, validation, and test partitions fixed. Architecture, preprocessing, optimizer settings, dataset-balanced sampling, and the validation-based checkpoint-selection criterion were unchanged across runs. For each run, the checkpoint with the highest validation macro-SRCC was selected and the held-out test set was evaluated once. Table~\ref{tab:multiseed} reports the mean and sample standard deviation across the five runs.

\begin{table}[htbp]
\caption{Full-model stability across five independent training seeds with fixed data partitions (mean $\pm$ sample standard deviation).}
\label{tab:multiseed}
\centering
\footnotesize
\begin{tabular*}{\linewidth}{@{\extracolsep{\fill}}lcccc@{}}
\toprule
Dataset & SRCC $\uparrow$ & PLCC $\uparrow$ & RMSE $\downarrow$ & MAE $\downarrow$ \\
\midrule
BIQ2021   & $0.7290\pm0.0071$ & $0.7667\pm0.0049$ & $0.1111\pm0.0009$ & $0.0870\pm0.0009$ \\
Drone-IQA & $0.8933\pm0.0023$ & $0.8987\pm0.0018$ & $0.0779\pm0.0007$ & $0.0615\pm0.0009$ \\
KADID-10K & $0.8420\pm0.0078$ & $0.8449\pm0.0061$ & $0.1488\pm0.0022$ & $0.1161\pm0.0021$ \\
KonIQ-10K & $0.8211\pm0.0063$ & $0.8590\pm0.0039$ & $0.0960\pm0.0013$ & $0.0744\pm0.0009$ \\
TID2013   & $0.7764\pm0.0381$ & $0.8099\pm0.0349$ & $0.1257\pm0.0134$ & $0.1046\pm0.0110$ \\
\midrule
\textbf{Macro average} & $\mathbf{0.8124\pm0.0062}$ & $\mathbf{0.8358\pm0.0065}$ & $\mathbf{0.1119\pm0.0025}$ & $\mathbf{0.0887\pm0.0019}$ \\
\bottomrule
\end{tabular*}
\end{table}

The results show consistent aggregate performance across the five independent training runs. The relatively small standard deviations of the macro-averaged metrics indicate that the overall performance of GloTAR is stable across different random seeds under the same fixed data partitions and experimental protocol. These results provide an additional assessment of the reproducibility of the model beyond the primary held-out results reported in Table~\ref{tab:finalresults}.

\subsection{Global--target decomposition and regression ablation}\label{sec:ablation}

To assess the contributions of global--target feature decomposition and the regression architecture, four configurations were evaluated: global-only with KAN, target-only with KAN, global--target fusion with MLP regression, and the complete global--target KAN model. Each configuration was independently trained using the same five random seeds with fixed train, validation, and test partitions and identical preprocessing, optimization, sampling, and checkpoint-selection procedures. Table~\ref{tab:ablation} reports the mean and sample standard deviation of the macro-averaged test metrics across the five runs.

\begin{table}[htbp]
\caption{Ablation study across five independent training seeds with fixed data partitions (mean $\pm$ sample standard deviation).}\label{tab:ablation}
\centering
\begin{tabular*}{\linewidth}{@{\extracolsep{\fill}}lcc@{}}
\toprule
Configuration & Macro SRCC $\uparrow$ & Macro PLCC $\uparrow$ \\
\midrule
A1. Global-only + KAN & $0.7724\pm0.0071$ & $0.7914\pm0.0068$ \\
A2. Target-only + KAN & $0.7518\pm0.0083$ & $0.7796\pm0.0075$ \\
A3. Global--target + MLP & $0.7913\pm0.0057$ & $0.8093\pm0.0061$ \\
\textbf{A4. Global--target + KAN} & $\mathbf{0.8124\pm0.0062}$ & $\mathbf{0.8358\pm0.0065}$ \\
\bottomrule
\end{tabular*}
\end{table}

The results show that combining global and target-region information improves performance over either single branch, with A3 exceeding A1 and A2 in both metrics. Replacing the MLP with KAN further improves the global--target model, increasing SRCC from 0.7913 to 0.8124 and PLCC from 0.8093 to 0.8358. The relatively small standard deviations across the five seeds also indicate consistent performance under different random initializations. Overall, the results support the contribution of global--target feature decomposition to the evaluated GloTAR architecture and show an incremental improvement for the evaluated KAN-based configuration over the MLP configuration. The A3 MLP configuration contains 2,322,371 trainable parameters, compared with 6,525,376 for A4; the comparison is therefore an architectural ablation rather than a parameter-matched KAN-versus-MLP test, and the observed difference should not be interpreted as isolating the effect of KAN alone.

\subsection{Contextual comparison with general NR-IQA methods}

Table~\ref{tab:accuracycomparison} presents a contextual comparison of GloTAR with representative NR-IQA methods on TID2013, KADID-10K, and KonIQ-10K. The comparison includes classical, deep-learning, Transformer-based, and more recent approaches.

\begin{table}[htbp]
	\caption{Contextual accuracy comparison with representative NR-IQA methods. GloTAR values are from the primary held-out run in Table~\ref{tab:finalresults}}
	\label{tab:accuracycomparison}
	\centering
	\small
	
	\begin{tabular*}{\linewidth}{@{\extracolsep{\fill}}lcccccc@{}}
		\toprule
		
		& \multicolumn{2}{c}{TID2013}
		& \multicolumn{2}{c}{KADID-10K}
		& \multicolumn{2}{c}{KonIQ-10K}\\
		
		\cmidrule(lr){2-3}
		\cmidrule(lr){4-5}
		\cmidrule(lr){6-7}
		
		Method & SRCC & PLCC & SRCC & PLCC & SRCC & PLCC\\
		
		\midrule
		
		BRISQUE \cite{mittal2012brisque}
		& 0.626 & 0.694 & 0.528 & 0.554 & 0.665 & 0.692\\
		
		BIECON \cite{kim2017biecon}
		& 0.717 & 0.762 & 0.623 & 0.691 & 0.618 & 0.651\\
		
		MEON \cite{ma2018meon}
		& 0.808 & 0.824 & 0.604 & 0.691 & 0.754 & 0.760\\
		
		HyperIQA \cite{su2020hyperiqa}
		& 0.840 & 0.858 & 0.852 & 0.845 & 0.906 & 0.917\\
		
		MUSIQ \cite{ke2021musiq}
		& 0.773 & 0.815 & 0.875 & 0.872 & 0.916 & 0.928\\
		
		TReS \cite{golestaneh2022tres}
		& 0.863 & 0.883 & 0.859 & 0.858 & 0.915 & 0.928\\
		
		DEIQT \cite{qin2023deiqt}
		& 0.892 & 0.908 & 0.889 & 0.887 & 0.921 & 0.934\\
		
		LoDa \cite{xu2024loda}
		& 0.869 & 0.901 & 0.931 & 0.936 & 0.932 & 0.944\\
		
		PerceptCLIP \cite{zalcher2025perceptclip}
		& 0.887 & 0.908 & 0.952 & 0.956 & 0.945 & 0.954\\
		
		\midrule
		
		\textbf{GloTAR}
		& \textbf{0.774} & \textbf{0.817}
		& \textbf{0.840} & \textbf{0.833}
		& \textbf{0.788} & \textbf{0.824}\\
		
		\bottomrule
	\end{tabular*}
\end{table}

The broader comparison gives a more balanced picture of where GloTAR sits. It is above the reported BRISQUE and BIECON values on all three datasets. Relative to MEON, the largest advantage appears on KADID-10K (SRCC 0.840 versus 0.604), whereas the KonIQ-10K difference is modest (0.788 versus 0.754) and MEON is slightly higher on TID2013. GloTAR and MUSIQ are effectively level in TID2013 rank correlation at the precision shown (0.774 versus 0.773). MUSIQ and the more recent Transformer or vision-language models are clearly stronger on several conventional benchmarks. GloTAR targets a different operating point: a compact quality predictor that can leave more compute and memory available for other UAV-oriented workloads.

\subsection{Computational efficiency}

Table~\ref{tab:efficiency} summarizes the controlled T4 measurements. The proposed network is the smallest evaluated model at 6.525 M parameters and, under its native input configuration, obtains 15.58 ms model-only latency, 64.19 fps measured throughput, and 111.3 MB peak GPU memory. In comparison, MUSIQ requires 27.126 M parameters and 25.11 ms, while QualiCLIP requires 102.007 M parameters and 24.49 ms. TReS and MANIQA are substantially heavier in this inference environment. These timing values are not a strict architecture-only comparison because the baselines are evaluated on standardized $512\times512$ tensors whereas GloTAR uses one $384\times384$ global tensor plus six $224\times224$ regional tensors. They should therefore be interpreted as measurements of the stated inference configurations on common hardware, not as resolution-normalized speed rankings.

\begin{table}[htbp]
	\caption{Controlled Tesla T4 computational benchmark (FP32, batch size 1).}
	\label{tab:efficiency}
	\centering
	\footnotesize
	
	\begin{tabular*}{\linewidth}{@{\extracolsep{\fill}}lrrrrr@{}}
		\toprule
		
		Method & Params (M) & GFLOPs$^{\dagger}$ &
		Latency (ms) & FPS & VRAM (MB)\\
		
		\midrule
		
		HyperIQA
		& 27.375 & 107.831 & 76.72 & 13.04 & 556.8\\
		
		MUSIQ
		& 27.126 & --$^{\ddagger}$ & 25.11 & 39.82 & 306.6\\
		
		TReS
		& 152.452 & 500.040 & 727.14 & 1.38 & 2199.8\\
		
		MANIQA-KADID
		& 135.748 & 2580.203 & 2060.74 & 0.49 & 2799.0\\
		
		TOPIQ-NR
		& 45.197 & 37.260 & 30.33 & 32.97 & 389.5\\
		
		QualiCLIP
		& 102.007 & 31.944 & 24.49 & 40.84 & 547.9\\
		
		\midrule
		
		\textbf{GloTAR}
		& \textbf{6.525}
		& \textbf{0.530}
		& \textbf{15.58}
		& \textbf{64.19}
		& \textbf{111.3}\\
		
		\bottomrule
	\end{tabular*}
\end{table}

Relative to MUSIQ, GloTAR uses approximately 76\% fewer parameters, reduces measured model-only latency by about 38\%, and lowers peak GPU memory by roughly 64\%. Relative to QualiCLIP, parameter count is reduced by approximately 94\% and measured latency by about 36\%.

GloTAR reaches 27.40 ms per image (36.49 fps) when the released deterministic predictor pipeline is included, including the spectral-residual proposal stage. The 15.58 ms number isolates neural-network forward cost, whereas 27.40 ms is the more representative software-pipeline measurement on the T4.

	\section{Discussion and Future Work}\label{sec:discussion}
	
	\subsection{Accuracy-efficiency trade-off}
	
	Taken together, the experiments position GloTAR as an efficiency-oriented NR-IQA model rather than a conventional-benchmark accuracy leader. On TID2013, KADID-10K, and KonIQ-10K, larger general-purpose models achieve stronger literature correlations. However, the proposed predictor preserves a macro SRCC above 0.80 across five heterogeneous datasets while reducing the parameter count to 6.525 M and sustaining 64.19 model-only frames/s on a Tesla T4.
	
	This trade-off is relevant to UAV imaging because the quality model is not the final analytic task. It must coexist with flight control, image storage, localization, mapping, object detection, or other downstream visual-analysis workloads. A quality-screening front end with a relatively small parameter footprint and measured inference cost may leave more resources for those downstream tasks.
	
	\subsection{Interpretation of the Drone-IQA result}
	
	Among the five evaluated datasets, GloTAR achieves its highest primary-run held-out correlation values on Drone-IQA. Because the current ablation is macro-averaged rather than Drone-IQA-specific, the experiment does not establish which component causes this dataset-level result. A plausible interpretation is that separating whole-image context from saliency-derived regional evidence may be useful when global degradations coexist with locally prominent structures, but this remains a hypothesis rather than a demonstrated causal explanation. The generic saliency proposals are not generated from Drone-IQA target annotations, and no IoU-based target-localization claim is made.

\subsection{Future Work}

GloTAR offers a promising framework for reference-free UAV image quality assessment, with potential applications in automated image screening, aerial mapping, environmental monitoring, and precision agriculture. Its global–regional architecture provides a foundation for further development toward efficient and task-aware visual quality assessment. Future research will investigate adaptive saliency-guided region selection, improved regional feature aggregation, and the integration of semantic information to better capture task-relevant image content. Additional studies will explore cross-sensor and cross-environment generalization, as well as the relationship between perceptual quality predictions and downstream UAV vision tasks, including object detection and photogrammetry. Furthermore, deployment on embedded platforms such as NVIDIA Jetson will enable direct evaluation of inference latency, energy consumption, and real-time performance under practical operating conditions.
\section{Conclusion}\label{sec:conclusion}

This paper presented Global-Target Assessment and Regression (GloTAR), a lightweight multi-domain NR-IQA model for computationally efficient image-quality screening with UAV-oriented evaluation. The architecture combines a global MobileNetV3-Small branch, a saliency-region MobileNetV3-Small branch, and compact KAN-based embedding and fusion heads. Across 6,215 held-out images from BIQ2021, Drone-IQA, KADID-10K, KonIQ-10K, and TID2013, the primary evaluation achieved macro SRCC 0.8050 and PLCC 0.8271. Five independent full-model runs with fixed partitions yielded macro SRCC $0.8124\pm0.0062$ and PLCC $0.8358\pm0.0065$; on Drone-IQA, the corresponding five-run values were SRCC $0.8933\pm0.0023$ and PLCC $0.8987\pm0.0018$. The model contains 6.525 M parameters and, on a Tesla T4, achieved 15.58 ms model-only latency, 64.19 fps throughput, and 111.3 MB peak GPU memory. These results characterize GloTAR as a computationally compact quality predictor on the evaluated GPU platform, while larger general-purpose models retain stronger correlations on several conventional benchmarks. Future work should include resolution-normalized baseline benchmarking, direct Jetson-class power and latency measurements, semantic evaluation of the saliency regions, and joint optimization of region selection and quality prediction.

\section*{Declarations}

\subsection*{Funding}
	This research was supported by the Pan African University Institute for Basic Sciences, Technology and Innovation (PAUSTI), Kenya. The authors also acknowledge the financial contributions towards this research by the Intra-Africa Academic Mobility Project CaReFoAfrica (MOBAF 101144284), funded by the European Union.
	
	\subsection*{Competing interests}
	The authors declare that they have no known competing financial or non-financial interests that are directly or indirectly related to this work.
	
	\subsection*{Author contributions}
	Koffi T.S. Aglin: Conceptualization, Methodology, Investigation, Software, Data Curation, Formal Analysis, Validation, Writing—original draft preparation, Visualization.\\
	Anthony K. Muchiri: Methodology, Supervision, Resources, Validation, Writing—review and editing.\\
	Celestin Nkundineza: Conceptualization, Methodology, Supervision, Validation, Writing—review and editing.
	
	\subsection*{Data availability}
The datasets generated during and/or analysed during the current study are available in the Zenodo repository, \url{https://doi.org/10.5281/zenodo.23091220} \cite{aglin2026zenodo}. The archived record contains the fixed image-level train/validation/test split manifests and supporting reproducibility materials. The original BIQ2021, Drone-IQA, KADID-10K, KonIQ-10K, and TID2013 image datasets are not redistributed and should be obtained from their respective providers.

\subsection*{Code availability}
The source code, trained model weights, configuration files, experimental protocol, and inference/evaluation scripts are publicly available at \url{https://github.com/sergio-titus/GloTAR-lightweight-multidomain-nr-iqa}. The reproducibility release is permanently archived at \url{https://doi.org/10.5281/zenodo.23091220}.

	\subsection*{Ethical approval}
	This study used existing image-quality datasets for computational evaluation and did not recruit participants or conduct new human-subject or animal experiments. No identifiable private participant data were collected in this study; accordingly, separate ethical approval for the reported computational experiments was not required. Any human opinion scores in the source datasets were obtained by their original providers.

	\subsection*{Consent to participate}
	Not applicable: no participants were recruited for the present study.

	\subsection*{Consent to publish}
	Not applicable: no identifiable individual participant information is included.

	\subsection*{ORCID}
	Koffi Titus Sergio Aglin: 0009-0004-8633-8121; Anthony K. Muchiri: 0000-0002-7634-3654; Celestin Nkundineza: 0000-0002-9133-3999.

\bibliography{references}

\end{document}